\documentclass[sigconf, table]{acmart}
\AtBeginDocument{%
  }
\AtBeginEnvironment{thebibliography}{\enlargethispage{4\baselineskip}\small}

\setcopyright{acmlicensed}
\copyrightyear{2026}
\acmYear{2026}
\acmDOI{10.1145/3805622.3810866}
\acmConference[ICMR '26]{2026 International Conference on Multimedia Retrieval}{June 16--July 3, 2026}{Chicago, IL, USA}
\acmISBN{979-8-4007-2617-0/2026/06}

\usepackage{xcolor}
\usepackage{siunitx}   
\usepackage{booktabs}  
\usepackage{multirow} 
\usepackage{float} 
\usepackage{subcaption}

\begin{document}

\title{CausalGS{}: Learning Physical Causality of 3D Dynamic Scenes with Gaussian Representations}



\author{Nengbo Lu}
\affiliation{%
 \institution{Guilin University of Electronic Technology}
 \city{Guilin}
 \country{China}}
\email{lunengbo@mails.guet.edu.cn}

\author{Minghua Pan}
\authornote{Corresponding author.}
\affiliation{%
  \institution{Guilin University of Electronic Technology}
  \city{Guilin}
  \country{China}
}
\email{panmh@guet.edu.cn}




\renewcommand{\shortauthors}{Trovato et al.}
\begin{abstract}
 Learning a physical model from video data that can comprehend physical laws and predict the future trajectories of objects is a formidable challenge in artificial intelligence. Prior approaches either leverage various Partial Differential Equations (PDEs) as soft constraints in the form of PINN losses, or integrate physics simulators into neural networks; however, they often rely on strong priors or high-quality geometry reconstruction. In this paper, we propose CausalGS, a framework that learns the causal dynamics of complex dynamic 3D scenes solely from multi-view videos, while dispensing with the reliance on explicit priors. At its core is an inverse physics inference module that decouples the complex dynamics problem from the video into the joint inference of two factors: the initial velocity field representing the scene's kinematics, and the intrinsic material properties governing its dynamics. This inferred physical information is then utilized within a differentiable physics simulator to guide the learning process in a physics-regularized manner. Extensive experiments demonstrate that CausalGS surpasses the state-of-the-art on the highly challenging task of long-term future frame extrapolation, while also exhibiting advanced performance in novel view interpolation. Crucially, our work shows that, without any human annotation, the model is able to learn the complex interactions between multiple physical properties and understand the causal relationships driving the scene's dynamic evolution, solely from visual observations. We anonymously provide the code at \url{https://github.com/DustSettled/CausalGS}.
\end{abstract}



\begin{teaserfigure}
  \includegraphics[width=0.95\textwidth]{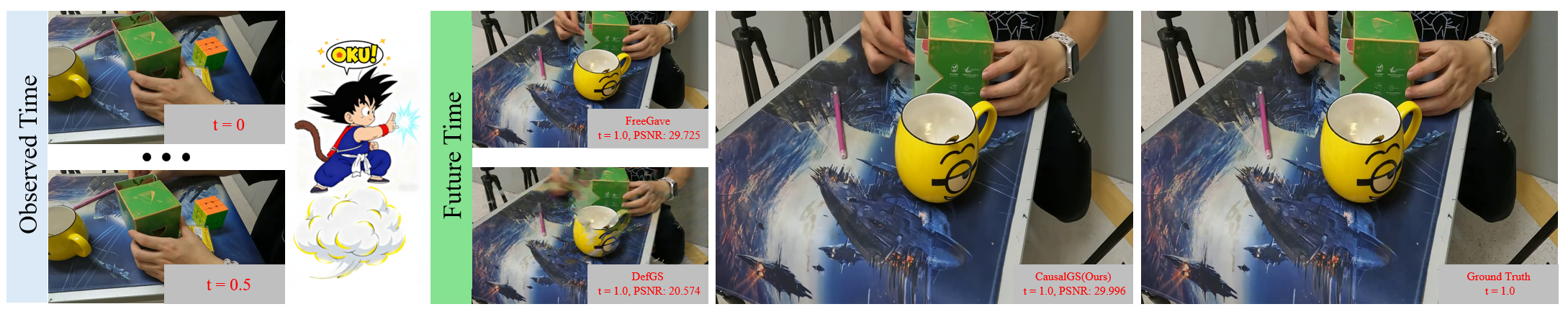}
  \caption{From a video stream of a real-world dynamic scene, CausalGS learns the interactions of physical properties within the scene, thereby understanding the causality that drives the scene's dynamic evolution and accurately predicting the future action of a box opening. Compared to existing methods, CausalGS demonstrates superior performance.}
  \Description{Overview teaser showing a real-world dynamic scene, the learned physical interactions, and a comparison in which CausalGS predicts a box-opening motion more accurately than prior methods.}\label{fig:1}
\end{teaserfigure}


\maketitle

\section{Introduction}
Constructing computational models capable of perceiving, understanding, and predicting the dynamic evolution of the physical world is a fundamental and long-standing goal. Acquiring accurate physical models\cite{lin2022diffskill,zheng2024physavatar,liu2024physics3d} of complex 3D environments is a key enabler for advancing cutting-edge applications such as robot planning, human-computer interaction\cite{xie2025generative,liu2021neural,xie2024physgaussian}, and dynamic content generation\cite{wu20244d,luiten2024dynamic,he2025survey,xie2025physanimator,lin2025omniphysgs}. In recent years, advances in neural scene representations, exemplified by Neural Radiance Fields (NeRF)\cite{shao2023tensor4d,liu2021neural,mildenhall2021nerf} and 3D Gaussian Splatting (3DGS)\cite{wu20244d,lin2024gaussian,xiang2024flashavatar}, have pushed the fidelity of visual reconstruction for dynamic 3D scenes to unprecedented levels, demonstrating excellent performance in tasks such as novel view interpolation and geometry reconstruction. However, these models essentially act as high-capacity function approximators for high-dimensional spatio-temporal data. They operate by learning the correlation between pixels and spatio-temporal coordinates, yet fail to capture the underlying causality that drives the scene's dynamic evolution. This paradigm, therefore, fundamentally hinders the effective learning of physical properties within the 3D space, leaving these models without the core capability for reliable future prediction.
For the learning of physical properties, existing works mainly follow two technical approaches. One is represented by PINNs, which embed Partial Differential Equations\cite{chu2022physics,park2023p} into the loss function as soft constraints. Although this paradigm has demonstrated potential in addressing continuum mechanics problems, it relies on spatiotemporal integration to impose constraints, leading to limited accuracy in physical identification when dealing with real-world scenarios involving complex boundaries or discontinuous contacts and often resulting in blurry future frame prediction.
The other approach uses an explicit and differentiable physical simulator\cite{cai2024gic,xie2024physgaussian,zhao2025synthetic,xiang2022dressing,bertiche2020pbns,su2023caphy} to provide strong physical priors for the learning process, enabling end-to-end gradient propagation from visual loss to physical parameters. Nevertheless, the applicability of such methods is usually confined to specific objects or motions, and they often rely on strong priors such as object masks, which severely restricts their scalability to unconstrained scenarios.

\begin{figure*}
    \centering
    \includegraphics[width=1\textwidth, height=5.5cm]{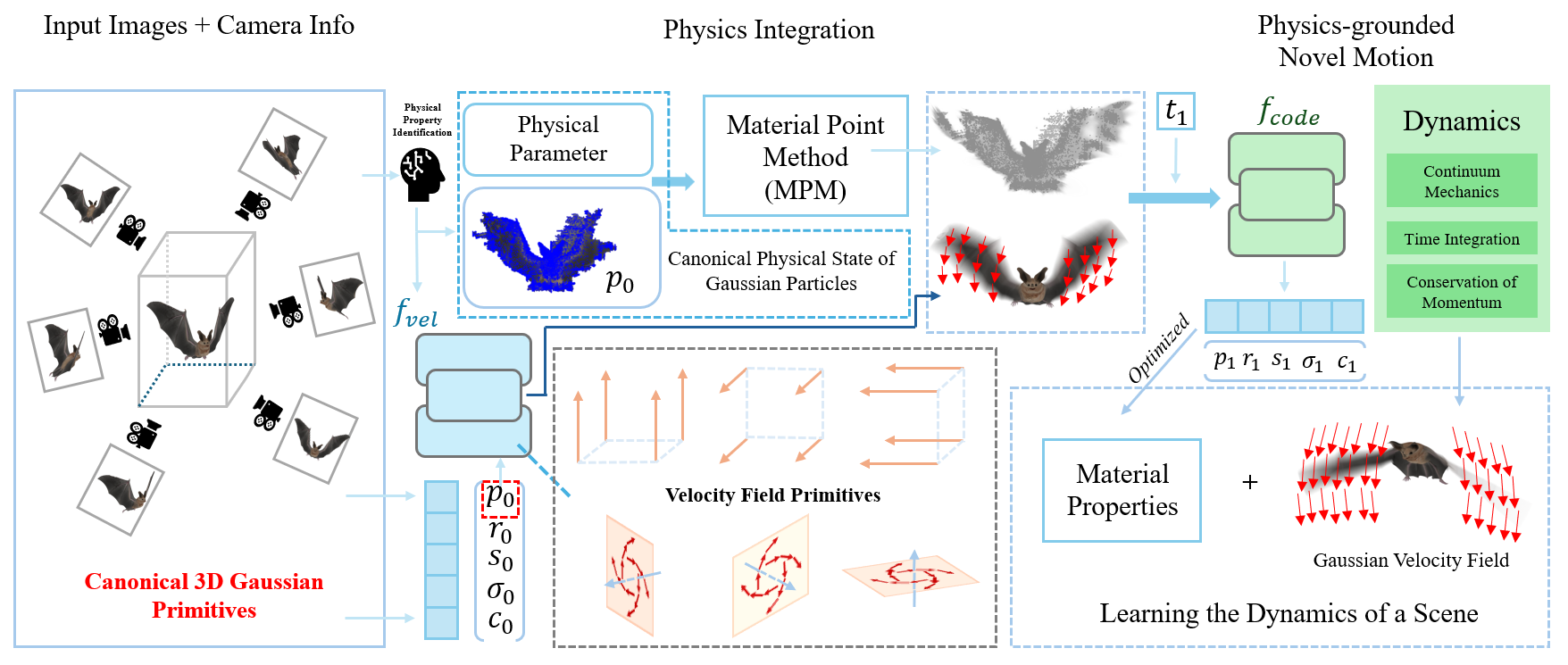}
    \caption{The overall pipeline of our model begins with multi-view image sequences of a scene from video, from which an inverse physics inference module identifies the scene's latent physical properties. The inferred properties then directly drive a differentiable physics simulator to evolve the scene's state according to physical laws. The model learns the interactions between the physical properties that drive the dynamic evolution of the scene. }
    \Description{Pipeline diagram showing multi-view video input, inverse physics inference, differentiable simulation, and the learned causal physical dynamics of the scene.}\label{fig:pipe}
\end{figure*}

In this work, our objective is to propose a novel framework for performing physical system identification of complex dynamic 3D scenes, sourcing directly from raw multi-view video streams as shown in Figure 1. Our approach dispenses with the reliance on PINN losses or the strong priors commonly necessitated by existing methods\cite{ma2023learning,song2024identifying,kamali2023elasticity}, such as object segmentation masks\cite{xiang20233d,yang2023reconstructing,xu2023human} or class labels\cite{zheng2024physavatar,guo2025pgc,liu2024clothedreamer,lin2025omniphysgs}, thereby enhancing its applicability in real-world scenarios. By learning 3D velocities and modeling complex material interactions, our framework acquires a physical model that is both more comprehensive in its representation and possesses greater capabilities for future prediction\cite{li2023nvfi,li2025freegave}.

Inferring authentic physical information from multi-view videos is a highly challenging inverse problem, the fundamental difficulty of which lies in the lack of physical and semantic constraints. In contrast to other approaches, we no longer attempt to model semantically complex objects\cite{ost2021neural,gao2024graphdreamer,ji2025jade}. Instead, we conceptualize the entire dynamic 3D scene as a continuous medium composed of a multitude of discretized physical particles. This particle-centric abstraction aligns naturally with our 3DGS representation, wherein each Gaussian primitive is treated as a physical particle\cite{wu20244d,luiten2024dynamic}. The collective dynamic behavior of this particle ensemble is not arbitrary but is instead governed by underlying physical principles. These principles can be decomposed into two core factors: the initial kinematic conditions and the intrinsic material properties.

Based on these core elements, we propose an efficient framework that jointly learns the 3D geometry, appearance, and underlying material properties of dynamic scenes. We treat each 3D Gaussian kernel as a physical particle and extend the existing 3D Gaussian Splatting (3DGS) method to equip the entire particle set with learnable initial dynamic conditions and intrinsic material properties. Leveraging these inferred physical parameters, a differentiable physical simulator performs forward simulation, and the resulting physical states provide additional supervision for learning the neural model. Specifically, our framework consists of three key components: a standard 3D representation module that adopts vanilla 3DGS to learn geometry, appearance, and material properties at canonical timestamps; an inverse physics inference module comprising a neural velocity network for regressing the initial velocity field and a material decoder for decoding intrinsic physical parameters; and a differentiable physical simulation and optimization module that takes the inferred physical properties as input, drives the physically realistic evolution of the scene, and enables end-to-end optimization via the final rendering loss.

The synergy between our modules is central to our framework, as shown in Figure~\ref{fig:pipe}. Given video frames of a dynamic 3D scene, the inverse physics inference module posits a set of hypotheses regarding the scene's physical properties. Subsequently, the differentiable physics simulation and optimization module simulates the scene's dynamic trajectory based on these hypotheses, explicitly driving the model to learn the physical properties that best explain the observed phenomena. This ensures that the entire predicted evolution of the scene adheres to complex physical laws, ultimately enabling precise predictions of future frames. In this paper, our main contributions are as follows:
\begin{itemize}
    \item We propose a general framework for learning the physics of dynamic 3D scenes solely from RGB videos, without the need for any object priors.
    \item We learn a complete physical model via physics simulation, which encompasses both initial velocities and intrinsic material properties.
    \item We demonstrate superior performance on the future frame extrapolation task across multiple real-world datasets featuring complex non-rigid interactions.
\end{itemize}

\section{Related Works}
\subsection{3D Representation Learning}
In recent years, both NeRF and 3DGS have emerged as landmark works in 3D scene representation learning. The core idea of the former is to utilize a MLP to map spatial coordinates and viewing directions to volume density and color, synthesizing high-quality novel-view images\cite{fridovich2023k,nam2023mip,wu2025swift4d,li2024st} via volume rendering. The latter, in contrast, represents the scene using a multitude of 3D Gaussian primitives, where each Gaussian explicitly defines its spatial position, 3D covariance, Spherical Harmonics (SH) coefficients, and opacity. Its rendering process discards the time-consuming ray marching inherent to NeRF and instead achieves real-time, photorealistic rendering through a differentiable rasterizer.
\subsection{Dynamic reconstruction}
To reconstruct dynamic scenes from monocular or multi-view videos and comprehend their physical underpinnings, prior works have leveraged static 3D representations for non-rigid reconstruction. These methods either reconstruct the scene in a frame-wise manner or maintain a canonical shape and model the deformation with a neural network\cite{wu20244d,yang2024deformable,wang2023co,li2025freegave,shao2023tensor4d}. While these methods have achieved impressive fidelity on the task of novel view synthesis within the observed timeframe, their capability to predict physically meaningful future states is fundamentally limited. This limitation stems from the fact that the learned deformation fields are essentially performing a form of kinematic interpolation. By overfitting to the visual observations, they fail to capture the causal dynamics that drive the scene's evolution. In our work, we learn the physical properties of each Gaussian from visual observations, guided by an inverse physics inference module and physics simulation.

\subsection{3D Physics Learning}
Inspired by differentiable physics simulation\cite{xie2024physgaussian,fan2023simulating,yu2023inferring,gao2025fluidnexus}, a line of subsequent methods performs inverse inference of physical parameters by comparing rendered results with 2D ground-truth images, often leveraging prior knowledge of the object's geometry. This enables the recovery of unknown geometry and physical properties of deformable objects from multi-view video sequences; however, the accuracy of the identified physics is tightly coupled with the quality of the geometry reconstruction. An alternative paradigm employs PINNs, which regularize the model to learn physical properties by translating physical laws into soft constraints within the loss function.
Our method, in contrast, achieves high-quality reconstruction within the observation time through dynamics-driven modeling. Concurrently, our inverse physics inference module identifies the material properties that drive a differentiable Material Point Method (MPM) simulator for learning.

\section{Method}
\subsection{Canonical 3D Gaussian Primitives}

\begin{table*}[htbp]
\centering
\caption{Quantitative evaluation of novel view interpolation and future frame extrapolation on the Dynamic Object and Dynamic Indoor Scene datasets\cite{li2023nvfi}. The best and second-best results are marked in pink and yellow, respectively.}\label{tab:dynamic_datasets}
\resizebox{\linewidth}{!}{
\label{tab:table1}
\begin{tabular}{lcccccccccccc}
\toprule
& \multicolumn{6}{c|}{Dynamic Object Dataset} & \multicolumn{6}{c}{Dynamic Indoor Scene Dataset} \\
\cmidrule{2-13}
& \multicolumn{3}{c|}{Interpolation} & \multicolumn{3}{c|}{Extrapolation} & \multicolumn{3}{c|}{Interpolation} & \multicolumn{3}{c}{Extrapolation} \\
\cmidrule{2-4} \cmidrule{5-7} \cmidrule{8-10} \cmidrule{11-13}
& PSNR$\uparrow$ & SSIM$\uparrow$ & LPIPS$\downarrow$ & PSNR$\uparrow$ & SSIM$\uparrow$ & LPIPS$\downarrow$ & PSNR$\uparrow$ & SSIM$\uparrow$ & LPIPS$\downarrow$ & PSNR$\uparrow$ & SSIM$\uparrow$ & LPIPS$\downarrow$ \\
\midrule
T-NeRF{\cite{pumarolad}}       & 13.163 & 0.709 & 0.353 & 13.818 & 0.739 & 0.324 & 24.944 & 0.742 & 0.336 & 22.242 & 0.700 & 0.363 \\
D-NeRF{\cite{pumarolad}}       & 14.158 & 0.697 & 0.352 & 14.660 & 0.737 & 0.312 & 25.380 & 0.746 & 0.300 & 20.791 & 0.692 & 0.369 \\
T-NeRF$_{\text{PINN}}$ & 15.286 & 0.794 & 0.293 & 16.189 & 0.835 & 0.230 & 26.250 & 0.461 & 0.638 & 23.290 & 0.477 & 0.414 \\
HexPlane$_{\text{PINN}}${\cite{cao2023hexplane}}  & 27.042 & 0.958 & 0.057 & 21.419 & 0.946 & 0.067 & 25.215 & 0.763 & 0.389 & 23.091 & 0.742 & 0.401 \\
NSFF{\cite{li2021neural}}         & - & - & - & - & - & - & 29.365 & 0.829 & 0.278 & 24.163 & 0.795 & 0.289 \\
TiNeuVox{\cite{fang2022fast}}     & 27.988 & 0.960 & 0.063 & 19.612 & 0.940 & 0.073 & 29.982 & 0.864 & 0.213 & 21.069 & 0.909 & 0.281 \\
NVFi{\cite{li2023nvfi}}         & 29.027 & 0.970 & 0.039 & 27.549 & 0.972 & 0.036 & 30.675 & 0.877 & 0.211 & 29.745 & 0.876 & 0.204 \\
DefGS{\cite{yang2024deformable}}        & 37.865 & 0.994 & 0.007 & 19.894 & 0.949 & 0.045 & 30.920 & 0.916 & 0.130 & 21.380 & 0.819 & 0.188 \\
DefGS$_{\text{nvfi}}$ & 37.316 & 0.994 & 0.008 & 28.749 & 0.984 & 0.013 & 29.176 & 0.915 & 0.133 & 31.096 & 0.945 & 0.077 \\
TRACE {\cite{li2025tracelearning3dgaussian}}      & - & - & - & 31.597 & 0.987 & 0.009 & - & - & - & 34.824 & 0.965 & 0.054 \\
FreeGave {\cite{li2025freegave}}      & \cellcolor{yellow!60}39.393 & \cellcolor{yellow!60}0.995 & \cellcolor{yellow!60}0.005 & \cellcolor{yellow!60}31.987 & \cellcolor{yellow!60}0.990 & \cellcolor{yellow!60}0.007 & \cellcolor{yellow!60}32.287 & \cellcolor{yellow!60}0.930 & \cellcolor{yellow!60}0.092 & \cellcolor{yellow!60}35.019 & \cellcolor{yellow!60}0.966 & \cellcolor{yellow!60}0.051 \\
\textbf{CausalGS (Ours)} & \cellcolor{pink!80}40.002 & \cellcolor{pink!80}0.995 & \cellcolor{pink!80}0.005 & \cellcolor{pink!80}34.517 & \cellcolor{pink!80}0.990 & \cellcolor{pink!80}0.007 & \cellcolor{pink!80}33.888 & \cellcolor{pink!80}0.930 & \cellcolor{pink!80}0.092 & \cellcolor{pink!80}36.748 & \cellcolor{pink!80}0.966 & \cellcolor{pink!80}0.050 \\
\bottomrule
\end{tabular}
}
\end{table*}

Our method adopts 3D Gaussian distributions as an explicit 3D scene representation, which exists in the form of a point cloud. We define $t=0$ as the canonical timestamp. Each 3D Gaussian distribution $G(x)$ for the variable $x$ is characterized by a covariance matrix $\Sigma$ and a center point $\mathbf{p}$, where $\mathbf{p}$ is referred to as the Gaussian mean.
\begin{equation}
G\!\left(\mathbf{x}\right) = \exp\left(-\frac{1}{2}{(\mathbf{x}-\mathbf{p})}^T \Sigma^{-1}(\mathbf{x}-\mathbf{p})\right).
\end{equation}

To enable separate optimization of the parameters, the covariance matrix $\Sigma$ can be decomposed into a scaling matrix $\mathbf{S}$ and a rotation matrix $\mathbf{R}$:
\begin{equation}
\Sigma = \mathbf{R} \mathbf{S} \mathbf{S}^{T} \mathbf{R}^{T}.
\end{equation}

When rendering a new viewpoint, the system employs a differentiable rasterization technique to project the 3D Gaussian distributions onto the camera plane. Via the view transformation matrix $\mathbf{W}$ and the Jacobian matrix $\mathbf{J}$ of the affine approximation of the projection, the covariance matrix $\Sigma'$ in the camera coordinate system can be computed as follows:
\begin{equation}
\Sigma' = \mathbf{J} \mathbf{W} \Sigma \mathbf{W}^T \mathbf{J}^T.
\end{equation}

Therefore, in our canonical space, each 3D Gaussian contains the following optimizable parameters: a spatial coordinate $\mathbf{p} \in \mathbb{R}^3$, a color $\mathbf{c}$ defined by k-dimensional Spherical Harmonic (SH) coefficients, an opacity $\alpha \in \mathbb{R}$, quaternion rotation parameters $\mathbf{q} \in \mathbb{R}^4$, and a 3D scaling factor $\mathbf{s} \in \mathbb{R}^3$. For the pixel shading process, each Gaussian point computes its color and opacity values based on its radiance field expression. For the $N$ ordered Gaussian points that cover a given pixel, the color blending formula is:
\begin{equation}
C = \sum_{i \in N} c_i \alpha_i \prod_{j=1}^{i-1}(1 - \alpha_j).
\end{equation}

Here, $c_i$ and $\alpha_i$ represent the color and opacity computed for a point by the 3D Gaussian $G_i$, which are jointly determined by its covariance $\Sigma_i$, its optimizable per-point opacity, and its SH color coefficients. This set of optimizable parameters collectively forms the initial state of the scene at $t=0$ and serves as the canonical state for the 3D Gaussian primitives. Within this framework, each Gaussian primitive is treated not only as a rendering element but also as a physical particle that will participate in the subsequent physics simulation.

\subsection{Inverse Physics Inference}
This module aims to infer, without explicit priors, the initial conditions that drive the evolution of each Gaussian primitive from multi-view videos: its velocity and material properties. Prevailing approaches predominantly follow two routes: the first directly regresses the velocity of each Gaussian but relies on inefficient PINN losses, which require dense spatio-temporal sampling, to impose physical constraints. The second introduces a latent physics code to encode motion patterns;
\begin{figure*}[t]
    \centering
    \includegraphics[width=0.95\textwidth]{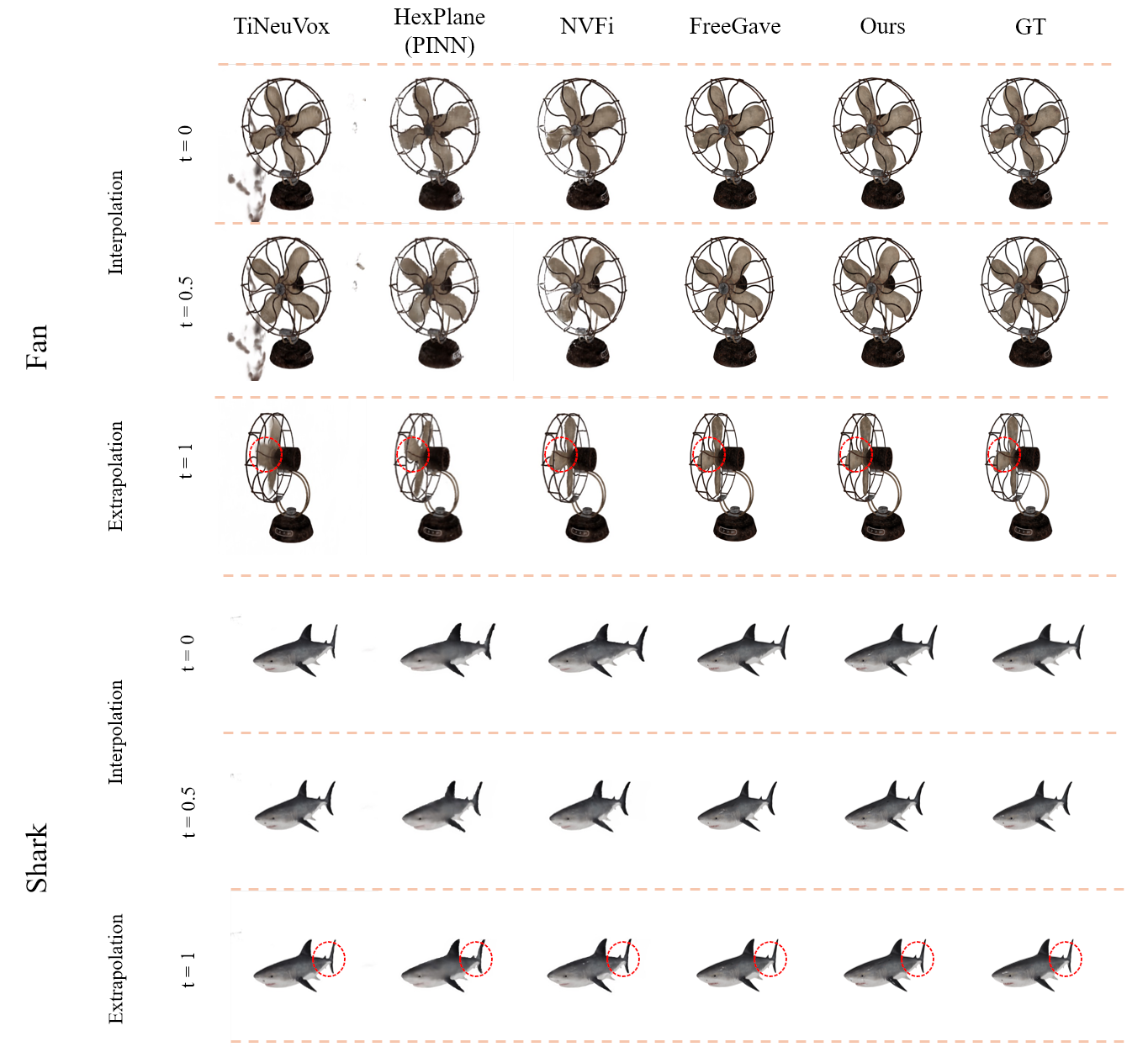}
    \caption{Qualitative comparison of our method against other models\cite{pumarolad,li2021neural,fang2022fast,li2023nvfi,yang2024deformable,li2025freegave} on interpolation and extrapolation tasks on the Dynamic Object Dataset.}
    \Description{Qualitative comparison on the Dynamic Object Dataset for interpolation and extrapolation, showing that our method preserves appearance and motion better than competing methods.}\label{fig:frame1}
\end{figure*}
however, this simplifies complex physical interactions into a single, entangled kinematic representation, neglecting the critical role of intrinsic dynamic factors such as material properties. This simplified modeling assumption inevitably leads to suboptimal performance on future prediction tasks that require causal reasoning. To address these limitations, we propose an inverse physics inference framework that jointly learns the velocity field and material properties in a self-supervised manner.

\begin{table}[htbp]
\centering
\caption{Quantitative results for novel view interpolation and future frame extrapolation on the FreeGave-GoPro dataset, rendered at a resolution of 960 × 540.}\label{tab:freegave_gopro}
\resizebox{\columnwidth}{!}{
\begin{tabular}{lcccccc}
\toprule
& \multicolumn{6}{c}{FreeGave-GoPro Dataset} \\
\cmidrule(lr){2-7}
& \multicolumn{3}{c|}{Interpolation} & \multicolumn{3}{c}{Extrapolation} \\
\cmidrule(lr){2-4} \cmidrule(lr){5-7}
 & PSNR$\uparrow$ & SSIM$\uparrow$ & LPIPS$\downarrow$ & PSNR$\uparrow$ & SSIM$\uparrow$ & LPIPS$\downarrow$ \\
\midrule
TiNeuVox {\cite{fang2022fast}} & 19.026 & 0.740 & 0.319 & 20.600 & 0.760 & 0.292 \\
NVFi {\cite{li2023nvfi}}     & 18.947 & 0.739 & 0.322 & 22.747 & 0.756 & 0.393 \\
DefGS {\cite{yang2024deformable}}    & 28.408 & 0.920 & 0.091 & 21.344 & 0.858 & 0.170 \\
DefGS$_{\text{nvfi}}$ & 28.217 & 0.920 & 0.092 & 26.431 & 0.903 & 0.124 \\
FreeGave {\cite{li2025freegave}}   & \cellcolor{yellow!60}28.451 & \cellcolor{yellow!60}0.920 & \cellcolor{yellow!60}0.091 & \cellcolor{yellow!60}28.094 & \cellcolor{yellow!60}0.914 & \cellcolor{yellow!60}0.112 \\
\textbf{CausalGS (Ours)}  & \cellcolor{pink!80}28.683 & \cellcolor{pink!80}0.920 & \cellcolor{pink!80}0.090 & \cellcolor{pink!80}28.267 & \cellcolor{pink!80}0.914 & \cellcolor{pink!80}0.111 \\
\bottomrule
\end{tabular}%
}
\end{table}

\noindent\textbf{Latent Physics Code.} The core idea of our method is to first learn a latent physics code $\mathbf{z}$, for each particle, rather than directly learning the velocity itself. This code is intended to provide a general description of the particle's intrinsic motion patterns and is shared across all timestamps, representing the particle's inherent tendency to move. We predict the physics code $\mathbf{z}_i$ for each canonical Gaussian from its position $\mathbf{p}_i$ using an MLP based network, $f_{vel}$:
\begin{equation}
\mathbf{z}_i = f_{vel}(\gamma(\mathbf{p}_i)),
\end{equation}
where $\gamma(\cdot)$ is a positional encoding function. This physics code, $\mathbf{z}_i$, is subsequently integrated into the representation of each canonical Gaussian primitive, serving as the key driver for its subsequent motion.

\noindent\textbf{Divergence-free Velocity Decomposition.} To ensure that the extrapolated future motion is physically plausible, the learned velocity field $\mathbf{v}(\mathbf{p}_t, t)$ must satisfy the fundamental divergence-free property. To this end, we treat each Gaussian as a rigid particle and decompose its velocity field into six basic velocity components, $\mathbf{V}_t$, and a basis matrix, $\mathbf{B}(\mathbf{p}_t)$, which depends only on the current position $\mathbf{p}_t$:
\begin{equation}
\mathbf{v}(\mathbf{p}_t, t) = \mathbf{V}_t \cdot \mathbf{B}(\mathbf{p}_t),
\end{equation}
where $\mathbf{V}_t \in \mathbb{R}^{1 \times 6}$ contains three linear and three angular velocity components. This formulation provides a strong inductive bias for structured motion while still allowing our material model to handle compressibility. The velocity component $\mathbf{V}_t$ is designed to be independent of the current position $\mathbf{p}_t$.

\noindent\textbf{Bottleneck Architecture for Velocity Components.} To learn the position-independent velocity component $\mathbf{V}_t$ from the physics code $\mathbf{z}_i$ and the timestamp $t$, we design a decoupled bottleneck architecture. Specifically, we first feed the physics code $\mathbf{z}_i$ into a network $f_{\text{neck}}$ to obtain a low-dimensional bottleneck vector $\mathbf{h}_i \in \mathbb{R}^K$. In parallel, we feed the timestamp $t$ into another network $f_{\text{weight}}$ to obtain a time-varying weight matrix $\mathbf{W}_t \in \mathbb{R}^{K \times 6}$. The final velocity component is then obtained by multiplying these two outputs:
\begin{equation}
\mathbf{V}_t = \mathbf{h}_i \cdot \mathbf{W}_t = f_{neck}(\mathbf{z}_i) \cdot f_{weight}(t).
\end{equation}
The intuition behind this design is that $f_{\text{neck}}$ is responsible for decoding the physics code into $K$ basic motion patterns, while $f_{\text{weight}}$ is responsible for generating a set of weights at each timestamp $t$ to linearly combine these motion patterns.

\noindent\textbf{Intrinsic Material Property Decoding.} Beyond kinematic information, the dynamic behavior of a scene is fundamentally determined by the intrinsic material $\mathbf{m}$ of its constituent objects. To this end, we introduce a material decoder, denoted $f_{\text{mat}}$, which is also implemented as a multilayer perceptron. This network infers the physical material parameters required by the MPM simulator for each particle, namely the Young's modulus $E_i$ characterizing elasticity and the Poisson's ratio $\nu_i$  characterizing incompressibility, among other parameters:
\begin{equation}
\mathbf{m}_i = (E_i, \nu_i, \dots) = f_{mat}(\gamma(\mathbf{p}_i)).
\end{equation}
In this manner, our model can learn the distinct physical properties of different parts of the scene.


\begin{table*}[htbp]
\centering
\caption{Quantitative results for novel view interpolation and future frame extrapolation on the NVIDIA Dynamic Scene dataset.}\label{tab:truck_skating}
\resizebox{\linewidth}{!}{
\begin{tabular}{lcccccccccccc}
\toprule
& \multicolumn{6}{c|}{Truck} & \multicolumn{6}{c}{Skating} \\
\cmidrule(lr){2-13}
& \multicolumn{3}{c|}{Interpolation} & \multicolumn{3}{c|}{Extrapolation} & \multicolumn{3}{c|}{Interpolation} & \multicolumn{3}{c}{Extrapolation} \\
\cmidrule(lr){2-4} \cmidrule(lr){5-7} \cmidrule(lr){8-10} \cmidrule(lr){11-13}
Model & PSNR$\uparrow$ & SSIM$\uparrow$ & LPIPS$\downarrow$ & PSNR$\uparrow$ & SSIM$\uparrow$ & LPIPS$\downarrow$ & PSNR$\uparrow$ & SSIM$\uparrow$ & LPIPS$\downarrow$ & PSNR$\uparrow$ & SSIM$\uparrow$ & LPIPS$\downarrow$ \\
\midrule
TiNeuVox{\cite{fang2022fast}}    & 27.230 & 0.846 & 0.229 & 24.887 & 0.848 & 0.209 & \cellcolor{pink!80}29.377 & 0.889 & 0.202 & 24.224 & 0.878 & 0.220 \\
HexPlane$_{\text{PINN}}${\cite{cao2023hexplane}} & 25.494 & 0.768 & 0.337 & 24.991 & 0.768 & 0.325 & 24.447 & 0.867 & 0.225 & 23.955 & 0.868 & 0.232 \\
NVFi{\cite{li2023nvfi}}          & 27.276 & 0.840 & 0.235 & 28.269 & 0.855 & 0.220 &26.999 & 0.848 & 0.227 & 28.654 & 0.896 & 0.208 \\
TRACE {\cite{li2025tracelearning3dgaussian}}      & 28.316 & 0.834 & 0.090 & 29.252 & 0.923 & 0.067 & 28.198 & 0.905 & 0.096 & \cellcolor{yellow!60}29.409 & 0.941 & 0.073 \\
FreeGave{\cite{li2025freegave}}          & \cellcolor{yellow!60}28.584 & \cellcolor{yellow!60}0.886 & \cellcolor{yellow!60}0.090 & \cellcolor{yellow!60}29.954 & \cellcolor{yellow!60}0.930 & \cellcolor{yellow!60}0.067 & 26.589 & \cellcolor{yellow!60}0.899 & \cellcolor{yellow!60}0.106 & 28.391 & \cellcolor{yellow!60}0.935 & \cellcolor{yellow!60}0.076 \\
\textbf{CausalGS (Ours)}   & \cellcolor{pink!80}28.924 & \cellcolor{pink!80}0.886 & \cellcolor{pink!80}0.090 & \cellcolor{pink!80}30.104 & \cellcolor{pink!80}0.931 & \cellcolor{pink!80}0.064 & \cellcolor{yellow!60}28.583 & \cellcolor{pink!80}0.902 &  \cellcolor{pink!80}0.010 & \cellcolor{pink!80}29.665 & \cellcolor{pink!80}0.935 & \cellcolor{pink!80}0.080 \\
\bottomrule
\end{tabular}%
}
\end{table*}

\subsection{Material Point Method}

After our inverse physics inference module successfully identifies the scene's initial dynamic conditions $\mathbf{V}$ and intrinsic material properties, we utilize a fully differentiable MPM physics simulator to forward-drive the dynamic evolution of the scene. We treat each 3D Gaussian primitive in the canonical space as a Lagrangian particle carrying its own physical state, with each particle representing a small material region as shown in Figure 4. Mass conservation in these Lagrangian particles ensures the constancy of total mass during movement. During each simulation time step $\Delta t$, the state evolution from time $n$ to $n+1$ begins by synchronizing the mass and momentum of each particle to the nodes $c$ of a background Eulerian grid to compute the grid node velocities $\mathbf{v}_c$.According to the conservation of momentum, the update of the grid node velocities is determined by the internal and external forces acting upon them.
\begin{equation}
m_i \frac{\mathbf{v}_c^{n+1} - \mathbf{v}_c^n}{\Delta t} = \mathbf{f}_c^{int} + \mathbf{f}_c^{ext}
\end{equation}
where $\mathbf{f}_c^{\text{ext}}$ is typically an external force. The internal force, $\mathbf{f}_c^{\text{int}}$, originates from the material's deformation and is obtained by mapping the stress tensor $\mathbf{P}_p$ of each particle back to the grid:
\begin{equation}
\mathbf{f}_c^{\text{int}} = -\sum_p V_p^0 \mathbf{P}_p {(\nabla w_{cp})}^T
\end{equation}
\begin{equation}
    \mathbf{v}_i^{n+1} = \sum_{c} w_{ic} \mathbf{v}_c^{n+1}
\end{equation}
\begin{equation}
\mathbf{p}_i^{n+1} = \mathbf{p}_i^n + \Delta t \cdot \mathbf{v}_i^{n+1}
\end{equation}
Here, the stress tensor $\mathbf{P}_p$ is a differentiable function, which is jointly determined by the particle's deformation gradient $\mathbf{F}_p$ and our inferred material properties $\mathbf{m}_i$. The gradient of the kernel function $w_{cp}$ with respect to the grid node position.
Subsequently, the updated grid node velocities $\mathbf{v}_c^{n+1}$ are interpolated back to each particle to update their velocities and deformation gradients. This process culminates in the final updated positions of the particles.

\begin{figure}
    \centering
    \includegraphics[width=0.47\textwidth]{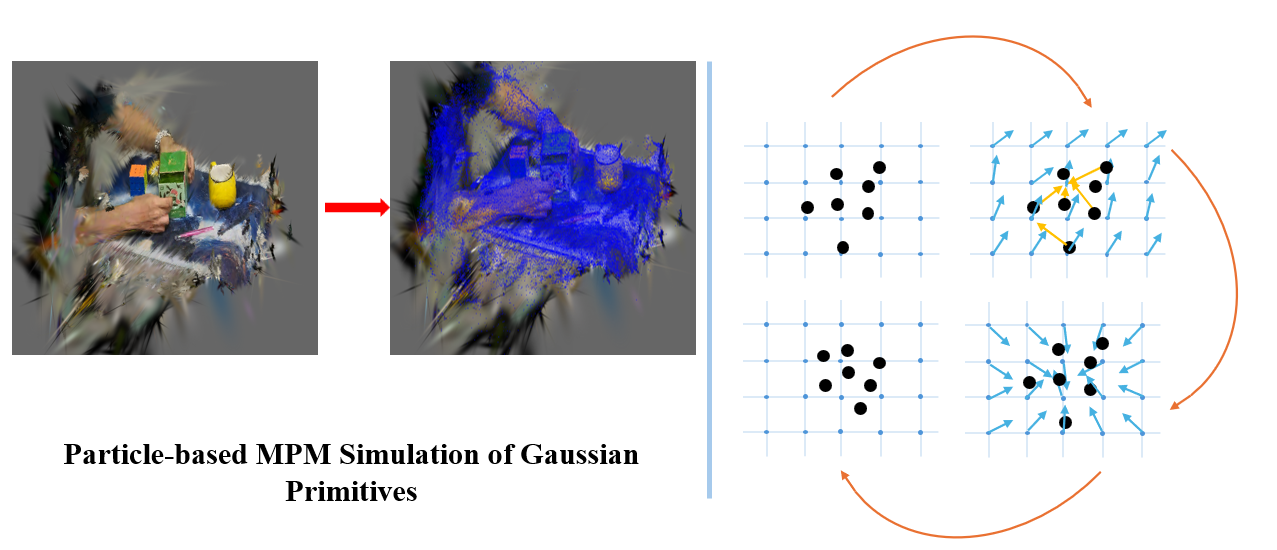}
    \caption{Each 3D Gaussian primitive is modeled as a Lagrangian particle and advanced with an MPM simulator.}
    \Description{Illustration of the particle-based simulation module, where Gaussian primitives become particles whose velocities and material properties are evolved by an MPM simulator.}\label{fig:frame}
\end{figure}



\begin{figure*}
    \centering
    \includegraphics[width=0.95\textwidth]{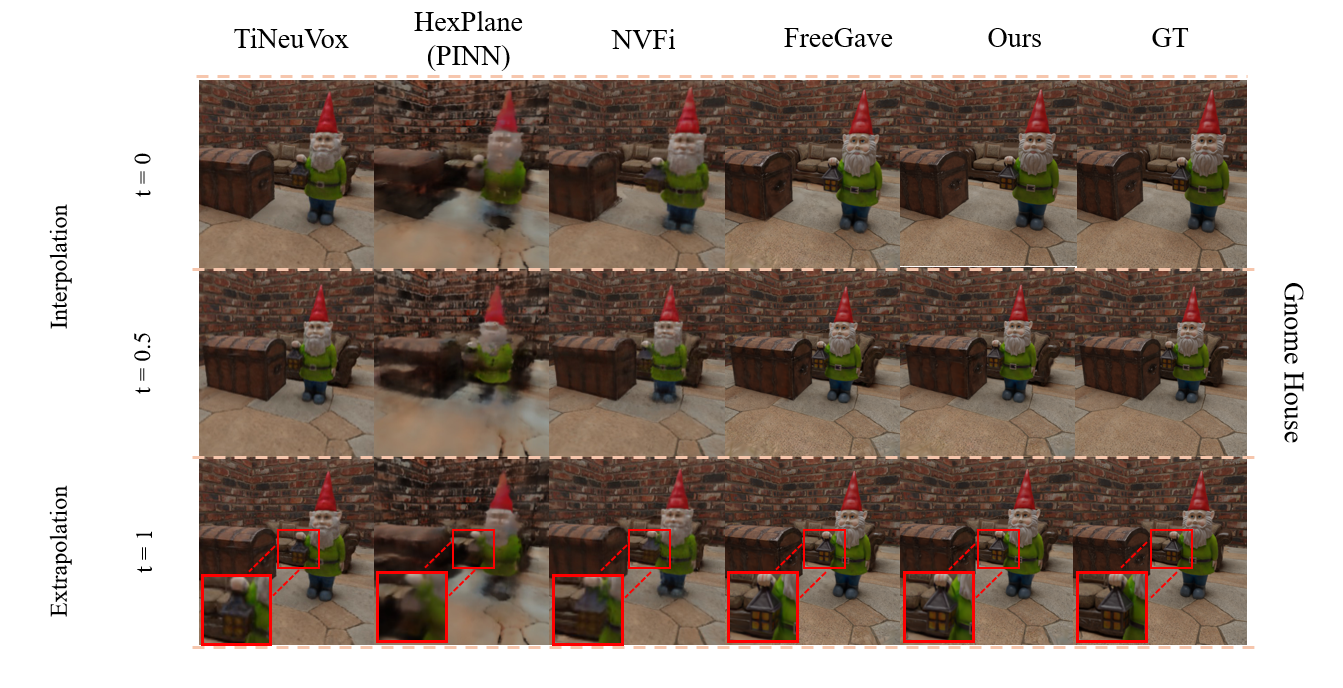}
    \caption{Qualitative comparison of rendering results against other models\cite{pumarolad,li2021neural,fang2022fast,li2023nvfi,yang2024deformable,li2025freegave} on the Dynamic Indoor Scene Dataset. We adopt the rendering settings from FreeGave\cite{li2025freegave}.}
    \Description{Qualitative rendering comparison on the Dynamic Indoor Scene Dataset, showing indoor motion and appearance reconstruction from our method and several baselines.}\label{fig:frame_indoor}
\end{figure*}

\section{Experiments}

\textbf{Datasets}: The core objective of our research is to learn a predictive physical model from dynamic videos, rather than simply performing interpolation fitting on observed training views. In line with this goal, we evaluate our method on two synthetic datasets previously used by the related work NVFi, as well as two real-world datasets.

The synthetic benchmarks are the \textbf{Dynamic Object Dataset}\cite{li2023nvfi}, which contains six objects with rigid or non-rigid deformations, and the \textbf{Dynamic Indoor Scene Dataset}\cite{li2023nvfi}, which contains four complex indoor scenes with challenging motions.

For real-world evaluation, we use two representative scenes from the public \textbf{NVIDIA Dynamic Scene Dataset}\cite{yoon2020novel}, captured by a static 12-camera rig. We also use the \textbf{FreeGave-GoPro Dataset}\cite{li2025freegave}, collected with a surrounding array of 20 GoPro Hero 10 Black cameras. All experiments were performed on a single NVIDIA RTX 3090 graphics processor with 24~GB of memory.

For each dynamic scene, we uniformly process 89 frames and resize the images to a resolution of 960 × 540. We use the first 67 frames from 17 of the views to train our model, while reserving the remaining 3 views within the same time period to evaluate the model's novel view interpolation capability. The final 22 frames from all 20 views are held out to assess future frame extrapolation, which serves as the primary validation for the predictive accuracy of our learned physical model.

\textbf{Baselines:} We have selected several methods for comparison. FreeGave stands as the most relevant work to ours, which primarily relies on constructing a divergence-free velocity field for each rigid particle. In contrast, our approach builds a physical model by learning a diverse set of physical properties for each particle. Existing methods such as T-NeRF\cite{pumarolad}, D-NeRF\cite{pumarolad}, NSFF\cite{li2021neural}, and TiNeuVox\cite{fang2022fast} are built upon the NeRF framework; however, they exhibit deficiencies in both reconstruction speed and rendering quality. Both FreeGave\cite{li2025freegave} and DefGS\cite{yang2024deformable} utilize the 3DGS framework and fit the motion trajectories of objects in the scene via a velocity field. Nevertheless, due to a lack of causal understanding of the scene's dynamics, they demonstrate suboptimal accuracy in future frame prediction and limited capabilities in novel view synthesis.

\textbf{Metrics:} For evaluating our method on the tasks of novel view interpolation and future frame extrapolation, we report the standard metrics PSNR, SSIM, and the LPIPS metric.

\subsection{Evaluation on Interpolation and Extrapolation}
We evaluate our method across four datasets~\cite{li2023nvfi,li2025freegave,yoon2020novel} on the two tasks of novel view interpolation and future frame extrapolation (Figure~\ref{fig:1}). Tables~\ref{tab:table1},~\ref{tab:freegave_gopro}, and~\ref{tab:truck_skating}, together with Figures~\ref{fig:frame1} and~\ref{fig:frame_indoor}, compare our approach with several baselines.

The experimental results demonstrate that our method achieves state-of-the-art performance across all datasets. It establishes a significant advantage on the highly challenging task of future frame extrapolation by outperforming all competing methods, and it also obtains the best results on the novel view interpolation task within the observed timeframe, notably surpassing strong baselines such as DefGS and FreeGave. This superior performance indicates that by learning 3D velocities and complex material interactions, CausalGS successfully grasps the causal dynamics of the scene, thereby enabling accurate predictions of future motions for single or multiple objects. Furthermore, it highlights the important role of physics simulation in helping the model understand scene motion while fully leveraging Gaussian primitives to represent both geometric structure and material properties.

\begin{figure*}
    \centering
    \includegraphics[width=0.95\textwidth]{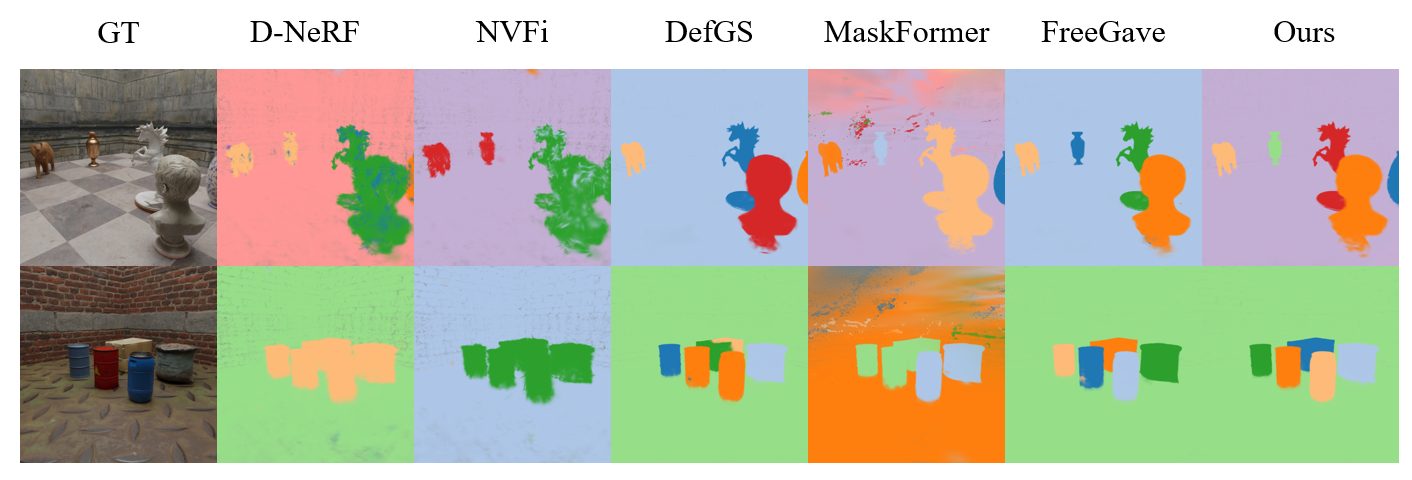}
    \caption{Qualitative results of our method for unsupervised motion segmentation.}
    \Description{Visualization of unsupervised motion segmentation, where the method groups scene regions according to learned physical motion patterns without semantic labels.}\label{fig:frame10}
\end{figure*}

\subsection{Analysis of Physics Code and Motion Patterns}
A core hypothesis of our work is that by learning the intrinsic physical properties of a scene, our model can understand the causal dynamics that drive its evolution. We posit that with this learned physical information, our model can spontaneously decompose the scene into distinct physical entities based solely on differences in motion patterns, all without any semantic labels. To validate this conjecture, we have designed the following experiment.

In our proposed inverse physics inference module, we learn the initial velocity and intrinsic material properties for each Gaussian particle. We reason that an ensemble of particles possessing similar physical properties and undergoing similar dynamic processes likely belongs to the same object or part in a semantic sense. Therefore, we select the intrinsic material property vector output by our material decoder, $f_{\text{mat}}$, as the physical signature for each particle.

\begin{table}[htbp]
\centering
\caption{Quantitative results of scene decomposition on the Synthetic Indoor Scene Dataset.}\label{tab:scene_decomposition}
\resizebox{\columnwidth}{!}{
\begin{tabular}{lcccccc}
\toprule
 & AP$\uparrow$ & PQ$\uparrow$ & F1$\uparrow$ & Pre$\uparrow$ & Rec$\uparrow$ & mIoU$\uparrow$ \\
\midrule
Mask2Former\cite{cheng2022masked} & 65.37 & 73.14 & 78.29 & 94.83 & 68.88 & 64.42 \\
D-NeRF\cite{pumarolad} & 57.26 & 46.15 & 59.02 & 56.55 & 62.94 & 46.58 \\
NVFi\cite{li2023nvfi} & 91.21 & 78.74 & 93.75 & 93.76 & 93.74 & 67.64 \\
DefGS\cite{yang2024deformable} & 51.73 & 57.60 & 66.43 & 63.21 & 70.07 & 54.46 \\
DefGS$_{nvfi}$\cite{yang2024deformable} & 55.26 & 62.75 & 69.83 & 69.39 & 72.91 & 56.82 \\
FreeGave\cite{li2025freegave} & \cellcolor{yellow!60}99.75 & \cellcolor{yellow!60}96.27 & \cellcolor{yellow!60}99.99 & \cellcolor{yellow!60}99.99 & \cellcolor{yellow!60}99.93 & \cellcolor{yellow!60}80.52 \\
\midrule
\textbf{CausalGS (Ours)} & \cellcolor{pink!80}99.82 & \cellcolor{pink!80}97.15 & \cellcolor{pink!80}99.99 & \cellcolor{pink!80}99.99 & \cellcolor{pink!80}99.93 & \cellcolor{pink!80}81.36 \\
\bottomrule
\end{tabular}%
}
\end{table}
For all Gaussian particles, we uniformly sample 10 timestamps to validate the learned physical information. Using the K-means clustering algorithm, we group them into $C$ clusters. We then assign a unique color to each cluster and render a segmentation map of the scene. As illustrated in Figure~5, we can clearly observe that this clustering, which is based purely on the learned physical properties, is remarkably capable of perfectly segmenting the different objects within the scene.

To quantitatively evaluate the model's emergent objectness, we conduct an unsupervised object segmentation benchmark on the Dynamic Indoor Scene dataset. We render the clustering results, which are based on physical properties, into 2D object segmentation masks and compare them against a series of representative baselines. These baselines include D-NeRF, NVFi, DefGS, and FreeGave. For the 3DGS-based baselines, we adopt a method that segments Gaussians based on the scene flow computed~\cite{song2022ogc,song2024unsupervised} from their learned deformation fields to ensure a fair comparison. Furthermore, we include a powerful, fully-supervised object segmentation model, Mask2Former~\cite{cheng2022masked}, pre-trained on the COCO dataset, as a reference baseline.
As shown in Table~\ref{tab:scene_decomposition} and Figure~\ref{fig:frame10}, our method outperforms all other baselines on all metrics.
We evaluate segmentation performance using AP, PQ, F1, Pre, Rec, and mIoU metrics to measure the consistency between our physics-based clustering and ground-truth objects.
\subsection{Ablation Study}
Our framework is composed of four core components: a 3D Gaussian primitive representation, an inverse physics inference module for identifying physical parameters, a velocity field for decoupling motion patterns, and a differentiable MPM simulator. We conduct the following ablation studies to evaluate the contribution of each component to the future frame extrapolation task on the FreeGave-GoPro dataset. The quantitative experimental results are shown in Table~\ref{tab:ablation}.

\textbf{Inverse Physics Inference (IPI).}
Our proposed Inverse Physics Inference Module transforms visual observations into the initial conditions required for physics simulation. The module comprises a neural velocity network to establish the initial velocity field and a material decoder to infer the intrinsic material properties of each particle. Unlike deformation fields that simply fit motion, we infer the underlying physical causes by disentangling the complex dynamics problem into an interaction of multiple physical factors.
\begin{table}[htbp]
\centering
\caption{Quantitative results of the ablation study on the FreeGave-GoPro Dataset. Here, ``K'' denotes the number of motion patterns.}\label{tab:ablation}
\setlength{\tabcolsep}{4pt}
\resizebox{\columnwidth}{!}{%
\begin{tabular}{cccccccc}
\toprule
\multirow{2}{*}{} & \multirow{2}{*}{$IPI$} & \multirow{2}{*}{$VFD$} & \multirow{2}{*}{$MPM$} & \multirow{2}{*}{\( K \)} & \multicolumn{3}{c}{Extrapolation} \\
\cmidrule(lr){6-8}
& & & & & PSNR$\uparrow$ & SSIM$\uparrow$ & LPIPS$\downarrow$ \\
\midrule
(1) & $\checkmark$ &  &  & 16 & 20.574 & 0.741 & 0.228 \\
(2) &  & $\checkmark$ &  &  16 & \cellcolor{yellow!60}27.921 & 0.911 &0.113 \\
(3) & $\checkmark$ & $\checkmark$ &  & 16 & 27.896 & \cellcolor{yellow!60}0.914 & \cellcolor{yellow!60}0.110 \\
(4) &  &  & $\checkmark$ & 16 & 22.032 & 0.874 & 0.141 \\
(5) & $\checkmark$ &  & $\checkmark$ & 16 & 23.905 & 0.881 & 0.139 \\
\midrule
\textbf{CausalGS (Ours)} & $\checkmark$ & $\checkmark$ & $\checkmark$ & 16 & \cellcolor{pink!80}28.267 & \cellcolor{pink!80}0.914 & \cellcolor{pink!80}0.110 \\
\bottomrule
\end{tabular}%
}
\end{table}
\textbf{Velocity Field Design (VFD).}
To validate the effectiveness of our velocity field design, we remove the core components of our method: the latent physics code and the bottleneck decomposition used for decoupling motion patterns. Instead, we employ a single, unstructured MLP that directly regresses the velocity from position and time.

\textbf{Material Point Method (MPM).}
We remove the MPM simulator and its associated physical regularization terms to validate the viability of employing a differentiable MPM simulator as the dynamics core. The experimental results in Table~\ref{tab:ablation} demonstrate that explicitly modeling material properties and contact dynamics reveals that the scene's dynamics are fundamentally linked to its dynamics, as represented by the material properties.

\section{Conclusion}
We extend 3D Gaussian Splatting with a framework for inverse physical inference from multi-view videos without strong priors. By leveraging an inverse inference module and a physics simulator, we endow Gaussian particles with learnable initial dynamics and intrinsic material properties to capture physical causality. Experiments across four datasets show superior performance in interpolation and extrapolation, illustrating a physics-driven paradigm for predicting the future in dynamic scene generation.

\begin{acks}
This work was supported by the National Natural Science Foundation of China (Grant No. 62361021) and the Guangxi Natural Science Foundation Project  (Grant No. 2025GXNSFAA069491).
\end{acks}
\bibliographystyle{ACM-Reference-Format}
\bibliography{sample-base}

\end{document}